\documentclass[10pt,twocolumn,letterpaper]{article}

\usepackage[pagenumbers]{cvpr} 

\usepackage{graphicx}
\usepackage{amsmath}
\usepackage{amssymb}
\usepackage{booktabs}
\usepackage{csquotes}
\usepackage{subcaption}

\definecolor{cvprblue}{rgb}{0.21,0.49,0.74}
\usepackage[pagebackref,breaklinks,colorlinks,allcolors=cvprblue]{hyperref}

\usepackage[capitalize]{cleveref}
\crefname{section}{Sec.}{Secs.}
\Crefname{section}{Section}{Sections}
\Crefname{table}{Table}{Tables}
\crefname{table}{Tab.}{Tabs.}

\def\paperID{*****}
\def\confName{CVPR}
\def\confYear{2026}

\begin{document}

\title{Gesture-Aware Pretraining and Token Fusion\\for 3D Hand Pose Estimation}

\author{Rui Hong, Jana Ko\v{s}eck\'{a} \\
George Mason University \\
{\tt\small \{rhong5, kosecka\}@gmu.edu}
}
\maketitle

\begin{abstract}
Estimating 3D hand pose from monocular RGB images is fundamental for applications
in AR/VR, human--computer interaction, and sign language understanding.
In this work we focus on a scenario where a discrete set of gesture labels is
available and show that gesture semantics can serve as a powerful inductive bias
for 3D pose estimation.
We present a two-stage framework: gesture-aware pretraining that learns an
informative embedding space using coarse and fine gesture labels from
InterHand2.6M~\cite{moon2020interhand2}, followed by a per-joint token
Transformer guided by gesture embeddings as intermediate representations
for final regression of MANO hand parameters~\cite{romero2017mano}.
Training is driven by a layered objective over parameters, joints, and structural
constraints.
Experiments on InterHand2.6M demonstrate that gesture-aware pretraining
consistently improves single-hand accuracy over the
state-of-the-art EANet~\cite{park2023extract} baseline, and that the benefit
transfers across architectures without any modification.
\end{abstract}

\section{Introduction}
\label{sec:intro}

Estimating 3D hand pose from monocular RGB images is a fundamental problem with
broad applications in augmented/virtual reality, human--computer interaction,
robotics, and sign language recognition. Despite rapid progress, the task remains
challenging due to severe self-occlusions, large articulation ranges, foreshortening,
and depth ambiguity.

Most existing approaches build upon the MANO parametric model~\cite{romero2017mano}
and regress hand pose and shape parameters from an RGB image, varying in their
intermediate representations, architectures, and learning strategies.
Early works, inspired by the success of SMPL for human mesh
recovery~\cite{kanazawa2018end}, directly regressed MANO parameters from
images~\cite{rong2021frankmocap,boukhayma20193d,zhou2020monocular,cai2018weakly}.
Follow-up methods improved performance using 2D keypoint heatmaps or larger-scale
datasets~\cite{ge20193d,chen2021model,zimmermann2019freihand}.
Structural constraints---bone length consistency, bone continuity, interpenetration
penalties---improved anatomical plausibility but do not address semantic
variation~\cite{cai2018weakly,yang2020bihand,zhou2020monocular}.
These advances exploit mostly geometric constraints and operate in a general
setting, rarely leveraging the \emph{semantic regularities} of hand gestures.
In many application domains, such as sign language, hand shapes correspond to
lexical units and exhibit recurring sub-configurations (e.g., characteristic
finger closures) that constrain feasible poses and aid disambiguation under
occlusion or low texture. Embedding such priors in the visual encoder can guide
feature formation and stabilize downstream structured reasoning.

\paragraph{Overview of our approach.}
We introduce a two-stage approach coupling gesture-aware pretraining with
gesture-guided fusion for robust 3D hand pose estimation.
In Stage~1, we pretrain an HRNet~\cite{wang2020deep} on InterHand2.6M
single-hand images with hierarchical gesture classification (coarse and fine
labels), endowing the encoder with semantics-sensitive features.
In Stage~2, the model computes per-joint tokens by integrating global image
features with a 2.5D volumetric representation and refines them via a
Transformer guided by gesture embeddings as intermediate representations to predict 3D hand pose.
Our contributions are:
\begin{itemize}
  \item A \textbf{gesture-aware pretraining} strategy that equips a
  high-resolution encoder with coarse-to-fine gesture semantics,
  motivated by the connection between hand pose and linguistic meaning in
  sign language~\cite{hosain2020finehand}.
  \item A \textbf{gesture-guided fusion framework} that uses gesture embeddings
  as intermediate representations to guide per-joint token refinement via a Transformer.
  \item Experimental validation showing that the proposed pretraining
  \textbf{transfers across architectures}: plugging our gesture-pretrained
  backbone into EANet~\cite{park2023extract} without any other modification
  consistently reduces error on InterHand2.6M.
\end{itemize}

\section{Related Work}

\paragraph{Single-Hand 3D Pose Estimation.}
Early approaches employed the MANO hand model~\cite{romero2017mano} and directly
regressed pose and shape parameters from an
image~\cite{boukhayma20193d,zhou2020monocular,cai2018weakly,rong2021frankmocap},
achieving compactness but struggling under occlusion.
Volumetric and 2.5D intermediate representations~\cite{ge20193d,zimmermann2019freihand,chen2021model}
provided stronger spatial supervision.
HaMeR~\cite{pavlakos2024hamer} and WiLoR~\cite{potamias2025wilor} further improved robustness
in challenging in-the-wild settings by pretraining on large mixed datasets including InterHand2.6M,
followed by fine-tuning on target benchmarks such as FreiHand~\cite{zimmermann2019freihand}.

\paragraph{Two-Hand Interactions.}
InterHand2.6M~\cite{moon2020interhand2} provided the first large-scale two-hand
dataset. Subsequent methods address the resulting severe mutual occlusion and
co-articulation via cross-hand attention~\cite{li2022interacting},
attentive implicit fields~\cite{lee2023im2hands}, extract-and-adapt
strategies~\cite{park2023extract}, and part-based attention~\cite{yu2023acr}.
Our work focuses on single-hand estimation; extending gesture semantics to
two-hand scenarios is a promising direction for future work.

\paragraph{Gesture Semantics.}
Gesture-aware representations have been exploited for sign language
recognition~\cite{koller2016deephand,hosain2020finehand,camgoz2017subunets,gueuwou-etal-2025-shubert},
but have not been utilized for 3D hand pose estimation.
Our work bridges this gap by explicitly integrating gesture-aware pretraining
into a pose estimation pipeline.

\section{Methodology}

The proposed approach consists of two stages:
(1) gesture-aware pretraining of an HRNet encoder with coarse- and fine-grained
gesture supervision, and
(2) per-joint tokenization and gesture-guided fusion for 3D pose estimation.
We adopt HRNet~\cite{wang2020deep} as the backbone, using its feature levels
$F_4$ and $F_5$ as multi-scale representations.

\subsection{Stage~1: Gesture-Aware Pretraining}

HRNet maintains high-resolution feature maps throughout the network by connecting
multi-resolution branches in parallel and repeatedly exchanging information across
them---a design proven effective for dense prediction tasks including human pose
estimation~\cite{sun2019deep}, hand pose estimation~\cite{potamias2025wilor},
object detection~\cite{wang2020deep}, and semantic
segmentation~\cite{sun2019high}.
We extend HRNet to hand pose estimation and equip it with gesture-aware pretraining
to inject semantic priors.

\begin{figure}[t]
  \centering
  \includegraphics[width=\linewidth]{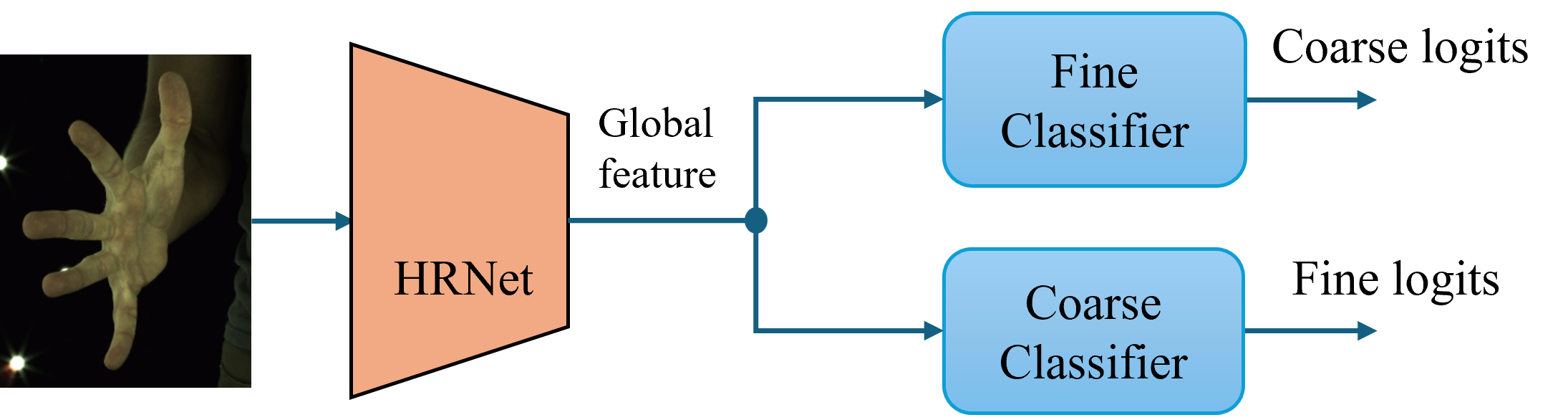}
  \caption{Stage~1: gesture-aware pretraining. HRNet global features $\mathbf{g}$
  feed two classification heads. A gradual weighting schedule stabilizes coarse
  categories first before incorporating fine distinctions.}
  \label{fig:stage1_pretrain}
\end{figure}

Among HRNet's feature levels, $F_4$ preserves higher-resolution spatial detail
while $F_5$ captures more semantic representations.
The global feature vector $\mathbf{g} \in \mathbb{R}^{512}$ is the pooled output
of $F_5 \!\in\! \mathbb{R}^{512\times16\times16}$:
\[
\mathbf{g} = \mathrm{Flatten}(\mathrm{Pool}(F_5)).
\]
Two lightweight classification heads predict gesture logits from $\mathbf{g}$:
\[
\boldsymbol{\gamma}_{\text{coarse}} = f_{\text{coarse}}(\mathbf{g}),
\quad
\boldsymbol{\gamma}_{\text{fine}} = f_{\text{fine}}(\mathbf{g}),
\]
and are trained with cross-entropy under a combined objective:
\[
L_{\text{cls}} = L_{\text{coarse}} + \alpha \, L_{\text{fine}}.
\]
To stabilize training, $\alpha$ is gradually increased from $0.1$ by $0.12$ every
10 epochs up to a maximum of $0.5$, prioritizing coarse semantic separation early
and progressively incorporating fine-grained cues.
An overview of Stage~1 is shown in \cref{fig:stage1_pretrain}.

After pretraining, the encoder produces multi-scale features $\{F_4, F_5\}$,
global feature $\mathbf{g}$, and gesture logits
$(\boldsymbol{\gamma}_{\text{coarse}}, \boldsymbol{\gamma}_{\text{fine}})$ reused in Stage~2.

\paragraph{Gesture label construction.}
InterHand2.6M contains 90 folders of single-hand gesture sequences.
We manually inspected images across folders and merged visually indistinguishable
gestures into a shared label, while assigning subtle variants to distinct
\emph{fine} labels under a common \emph{coarse} category.
This yielded 54 coarse and 70 fine gesture categories used as Stage~1 supervision.

\begin{figure}[t]
  \centering
  \includegraphics[width=\linewidth]{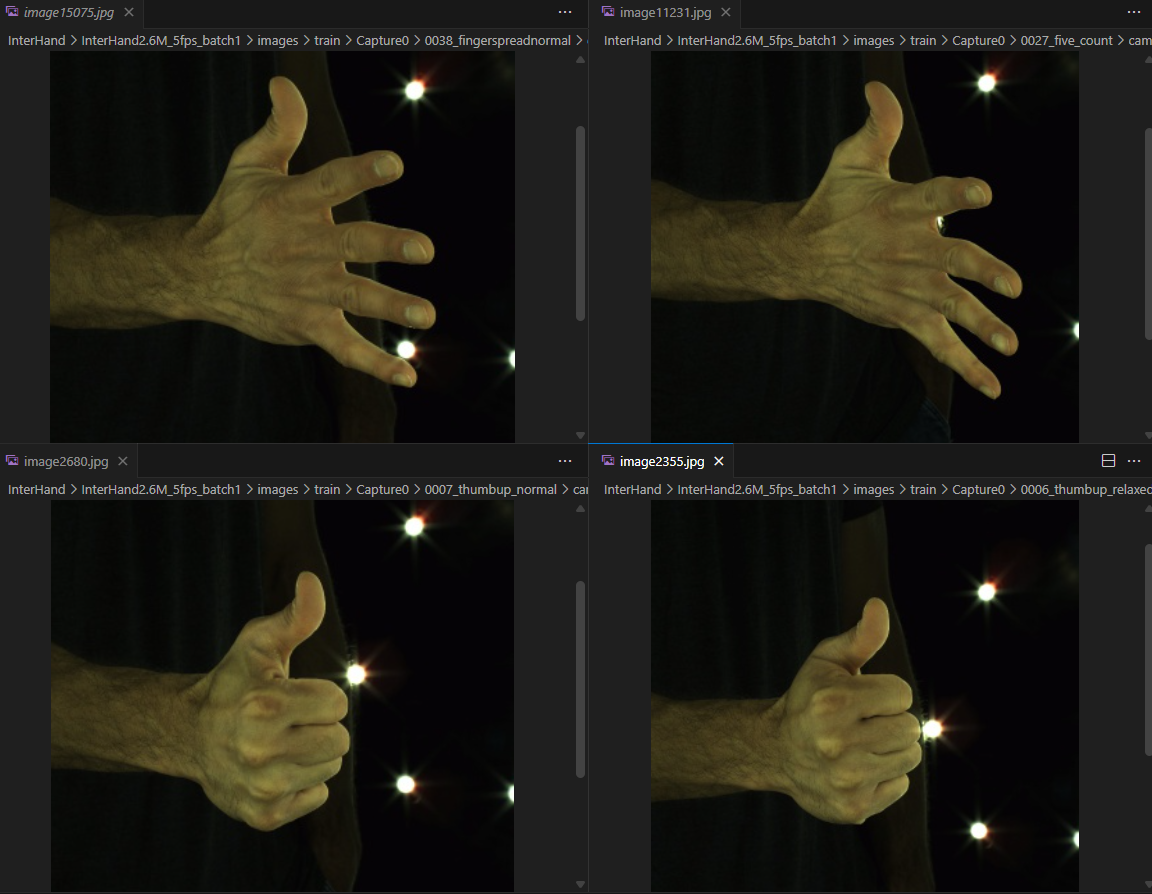}
  \caption{Example gesture images.
  Top: \enquote{fingerspreadnormal} and \enquote{five\_count} (same label).
  Bottom: \enquote{thumbup\_normal} and \enquote{thumbup\_relaxed}
  (same coarse, different fine label).}
  \label{fig:gesture_4_example}
\end{figure}

As shown in \cref{fig:gesture_4_example}, folders
\enquote{0038\_fingerspreadnormal} and \enquote{0027\_five\_count} are visually
identical and receive the same label at both levels.
Conversely, \enquote{0007\_thumbup\_normal} and \enquote{0006\_thumbup\_relaxed}
share a coarse label but differ in fine label due to a subtle thumb position
difference (\cref{tab:gesture_labels}).

\begin{table}[t]
\centering
\caption{Example gesture labeling. $c_i \in [0,54)$: coarse index;
$f_j \in [0,70)$: fine index.}
\begin{tabular}{c|c|c}
\hline
\textbf{Image} & \textbf{Coarse ($c_i$)} & \textbf{Fine ($f_j$)} \\
\hline
Top-left     & $c_{1}$ & $f_{1}$ \\
Top-right    & $c_{1}$ & $f_{1}$ \\
Bottom-left  & $c_{2}$ & $f_{2}$ \\
Bottom-right & $c_{2}$ & $f_{3}$ \\
\hline
\end{tabular}
\label{tab:gesture_labels}
\end{table}

\subsection{Stage~2: Gesture-Guided 3D Pose Estimation}

\begin{figure}[t]
\centering
\includegraphics[width=\linewidth]{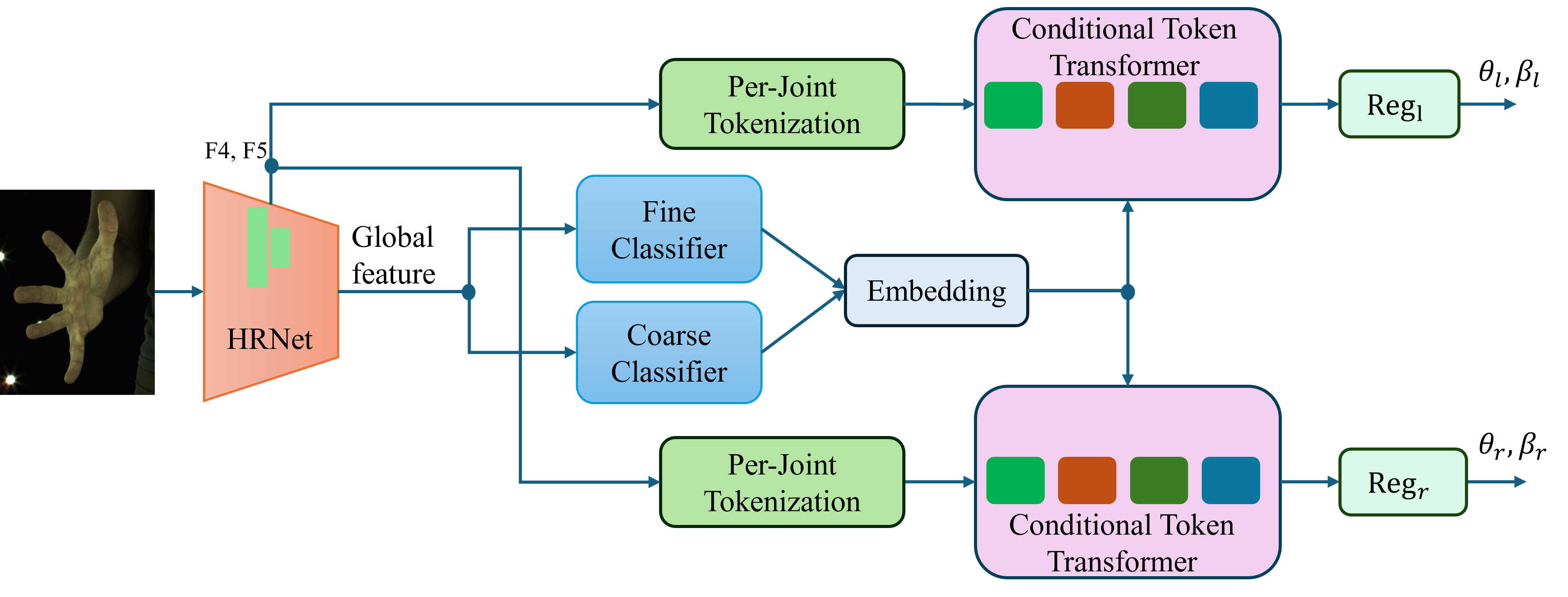}
\caption{Stage~2 pipeline. Multi-scale features ($F_4$, $F_5$) feed the
per-joint tokenization module. Stage~1 gesture logits
$(\boldsymbol{\gamma}_{\text{coarse}}, \boldsymbol{\gamma}_{\text{fine}})$ are embedded and injected into the gesture-guided Transformer.
Outputs are decoded to MANO parameters $(\theta, \beta)$.}
\label{fig:singlehand}
\end{figure}

Stage~2 constructs per-joint tokens from HRNet multi-scale features using a 2.5D
volumetric representation, then refines them via a Transformer guided by
Stage~1 gesture embeddings as intermediate representations.
The full pipeline is shown in \cref{fig:singlehand}.

From the pretrained HRNet encoder we take
$F_4\!\in\!\mathbb{R}^{256\times32\times32}$ and
$F_5\!\in\!\mathbb{R}^{512\times16\times16}$,
and the global descriptor $\mathbf{g}\!\in\!\mathbb{R}^{512}$.
The Stage~1 gesture logits $(\boldsymbol{\gamma}_{\text{coarse}}, \boldsymbol{\gamma}_{\text{fine}})$ are projected into the token embedding space via two
independent two-layer MLPs:
\[
\mathbf{e}_{\text{coarse}} = \phi_{\text{coarse}}(\boldsymbol{\gamma}_{\text{coarse}}), \quad
\mathbf{e}_{\text{fine}} = \phi_{\text{fine}}(\boldsymbol{\gamma}_{\text{fine}}),
\]
\[
\mathbf{e}_{\text{coarse}},\, \mathbf{e}_{\text{fine}} \in \mathbb{R}^{1024}.
\]

\paragraph{Per-Joint Tokenization via 2.5D Volumetric Reasoning}

$F_4$ and $F_5$ are each projected to $\mathbb{R}^{512\times16\times16}$ and concatenated along
channels to form $\mathbf{F}\!\in\!\mathbb{R}^{1024\times16\times16}$.
Flattening yields a global token sequence
$\mathbf{T}\!\in\!\mathbb{R}^{256\times1024}$,
which is passed through a lightweight Transformer encoder and reshaped to a
spatial feature map $\mathbf{F}'\!\in\!\mathbb{R}^{1024\times16\times16}$.

A volumetric head predicts a full 3D heatmap volume per joint,
$\mathbf{H}\!\in\!\mathbb{R}^{J\times16\times16\times16}$, $J{=}21$.
Joint locations are estimated via soft-argmax over this volume:
\[
(\hat{x},\hat{y},\hat{z})
= \sum_{x,y,z} (x,y,z)\,
\frac{\exp\!\big(\mathbf{H}(j,x,y,z)\big)}
     {\sum_{x',y',z'} \exp\!\big(\mathbf{H}(j,x',y',z')\big)},
\]
where $(\hat{x},\hat{y})$ lie in the image plane and $\hat{z}$ is a
wrist-relative depth offset rather than metric depth.
We refer to this as a \emph{2.5D} representation following~\cite{moon2020interhand2}:
the volume itself is 3D, but the $z$-axis encodes relative depth rather than
absolute 3D position, distinguishing it from full metric 3D reconstruction.
Using $(\hat{x},\hat{y})$, we sample $\mathbf{F}'$ via \texttt{grid\_sample}
to obtain per-joint tokens $\mathbf{U}\!\in\!\mathbb{R}^{1024\times J}$.

\paragraph{Transformer Module with Gesture Guidance Tokens}

The per-joint tokens are linearly projected and augmented with a learnable
positional encoding:
\[
\mathbf{Q} = \mathbf{U}^{\top} W + \mathbf{E}_{\text{pos}},
\quad \mathbf{Q}\in\mathbb{R}^{J\times 1024}.
\]
Gesture guidance tokens are formed as:
\[
\mathbf{c}_\star = \sigma(s_\star)\,\big(\mathbf{e}_\star + \mathbf{t}_\star\big),
\quad \star \in \{\text{coarse},\, \text{fine}\},
\]
where $\mathbf{t}_\star$ is a learnable type embedding and $s_\star$ is a scalar
gate initialized near zero for stable training.
The joint and gesture guidance tokens are concatenated:
\[
\mathbf{S} = \big[\,\mathbf{Q}\;;\;\mathbf{c}_{\text{coarse}}\;;\;\mathbf{c}_{\text{fine}}\,\big]
\in \mathbb{R}^{(J+2)\times 1024},
\]
and processed by a Transformer encoder, retaining only the first $J$ positions:
\[
\mathbf{O} = \big(\mathcal{E}(\mathbf{S})\big)_{1:J} \in \mathbb{R}^{J\times 1024}.
\]
A residual connection fuses refined tokens back with the original per-joint features:
\[
\widetilde{\mathbf{U}} = \mathbf{U} + \mathrm{Post}\!\big(\mathbf{O}^{\top}\big),
\qquad \widetilde{\mathbf{U}} \in \mathbb{R}^{1024\times J},
\]
where $\mathrm{Post}(\cdot)$ is a learnable $1{\times}1$ convolution.

\paragraph{MANO Parameter Regression}

From $\widetilde{\mathbf{U}}$, two MLP heads regress MANO parameters:
(i) a pose head predicting joint rotations in the 6D
representation~\cite{zhou2019continuity} for continuity and stability, and
(ii) a shape head regressing shape coefficients $\boldsymbol{\beta}\in\mathbb{R}^{10}$.

\paragraph{Training Objective}

The training loss combines three levels of supervision:
\begin{align}
\mathcal{L} ={}&
\lambda_{p}\, L_{\text{pose}}
+ \lambda_{s}\, L_{\text{shape}} \nonumber \\
&+ \lambda_{3\text{D}}\, L_{3\text{D}}
+ \lambda_{\text{mano-3D}}\, L_{\text{mano-3D}}
+ \lambda_{2.5\text{D}}\, L_{2.5\text{D}} \nonumber \\
&+ \lambda_{2\text{D}}\, L_{2\text{D}}
+ \lambda_{\text{mano-2D}}\, L_{\text{mano-2D}} \nonumber \\
&+ \lambda_{\text{cont}}\, L_{\text{cont}},
\end{align}
with weights $\lambda_{p}{=}2$, $\lambda_{s}{=}0.5$,
$\lambda_{3\text{D}}{=}\lambda_{\text{mano-3D}}{=}20$,
$\lambda_{2\text{D}}{=}\lambda_{\text{mano-2D}}{=}0.5$,
$\lambda_{2.5\text{D}}{=}0.05$, $\lambda_{\text{cont}}{=}10$.

\noindent\textbf{Parameter-level.}
$L_{\text{pose}}$ and $L_{\text{shape}}$ are $\ell_1$ losses on the predicted MANO
pose parameters (6D joint rotations) and shape coefficients $\boldsymbol{\beta}$,
respectively.

\noindent\textbf{Joint-level.}
$L_{3\text{D}}$ is an $\ell_1$ loss on camera-space 3D joint positions obtained
from the 2.5D soft-argmax predictions.
$L_{\text{mano-3D}}$ is an $\ell_1$ loss on the 3D joints produced by the MANO
forward kinematics layer from the regressed parameters.
$L_{2.5\text{D}}$ is an $\ell_1$ loss on the 2.5D heatmap coordinates
(image-plane location and wrist-relative depth) predicted by the volumetric head.
$L_{2\text{D}}$ and $L_{\text{mano-2D}}$ are the corresponding 2D reprojection
losses for the soft-argmax and MANO joints, respectively.

\noindent\textbf{Structural.}
$L_{\text{cont}}$ enforces bone continuity along kinematic chains by penalizing
discontinuities in consecutive joint rotations, encouraging anatomically smooth poses.

\section{Experiments}

We evaluate on the single-hand subset of
InterHand2.6M~\cite{moon2020interhand2} and report MPJPE and MPVPE in
millimeters. All methods in \cref{tab:single_hand_exp_results} are trained
on single-hand images of InterHand2.6M, except HaMeR which is pretrained on a
large mixed dataset without InterHand2.6M fine-tuning; its higher error reflects
this domain gap rather than a fundamental limitation of the approach.

\subsection{Gesture-Aware Pretraining Analysis}

\begin{figure}[t]
    \centering
    \begin{subfigure}{0.45\linewidth}
        \includegraphics[width=\linewidth]{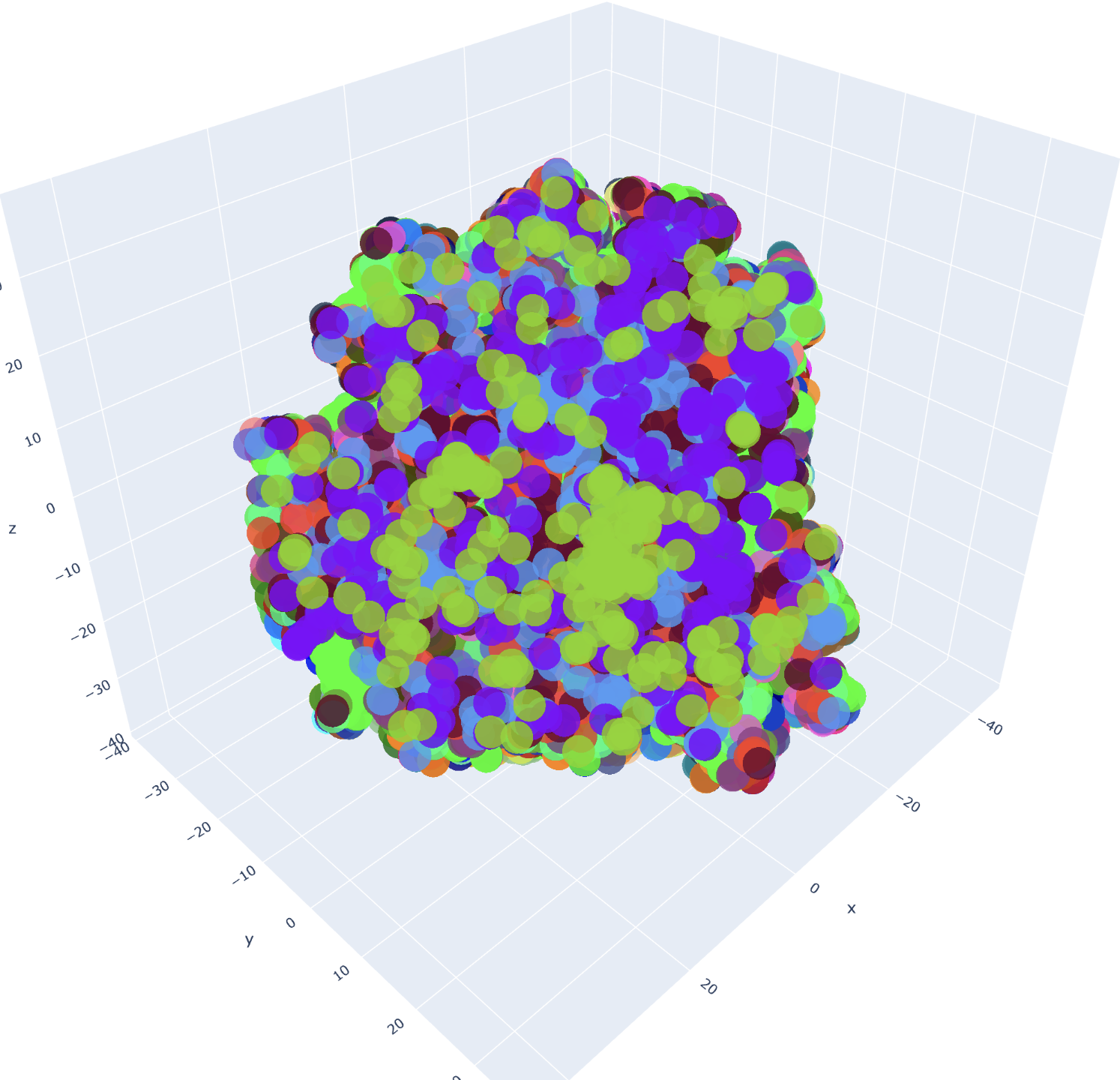}
        \subcaption{No pretraining}
    \end{subfigure}
    \hfill
    \begin{subfigure}{0.45\linewidth}
        \includegraphics[width=\linewidth]{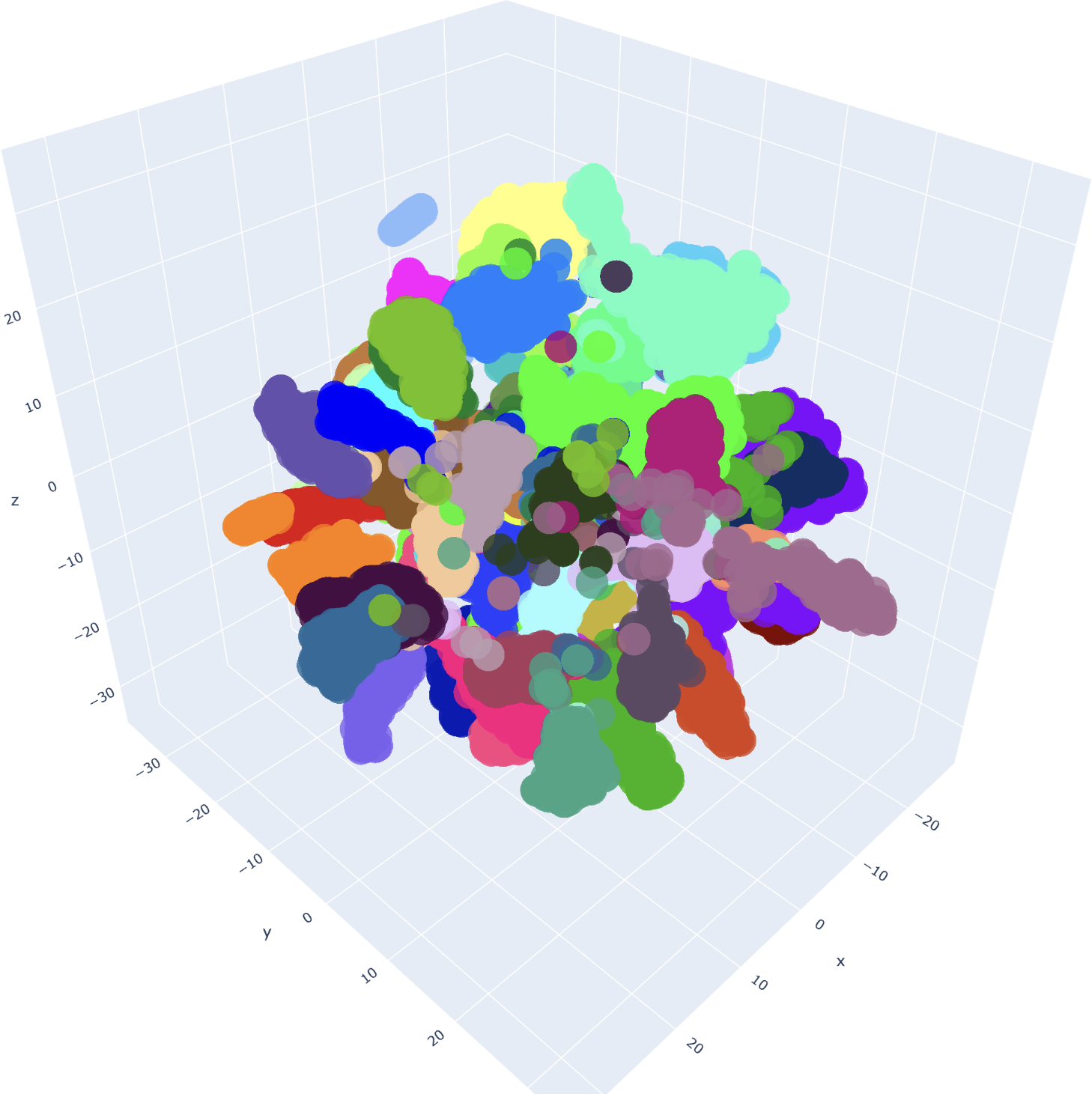}
        \subcaption{With pretraining}
    \end{subfigure}
    \caption{t-SNE of HRNet pooled features on InterHand2.6M test set, colored
    by coarse gesture label. Gesture pretraining yields more discriminative
    representations.}
    \label{fig:tsne_pool}
\end{figure}

\begin{figure}[t]
    \centering
    \begin{subfigure}{0.45\linewidth}
        \includegraphics[width=\linewidth]{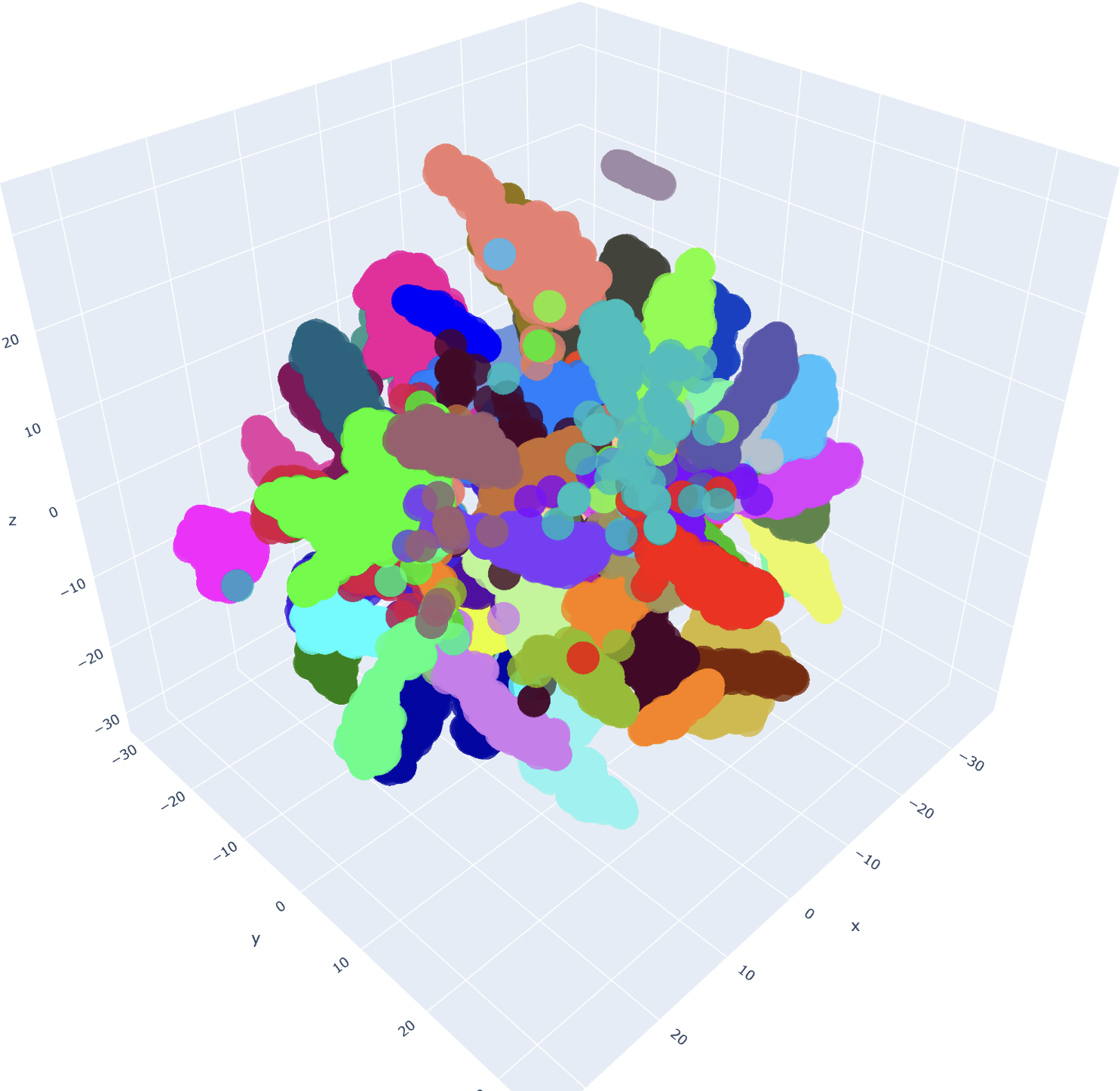}
        \subcaption{Coarse logits $\boldsymbol{\gamma}_{\text{coarse}}$}
    \end{subfigure}
    \hfill
    \begin{subfigure}{0.45\linewidth}
        \includegraphics[width=\linewidth]{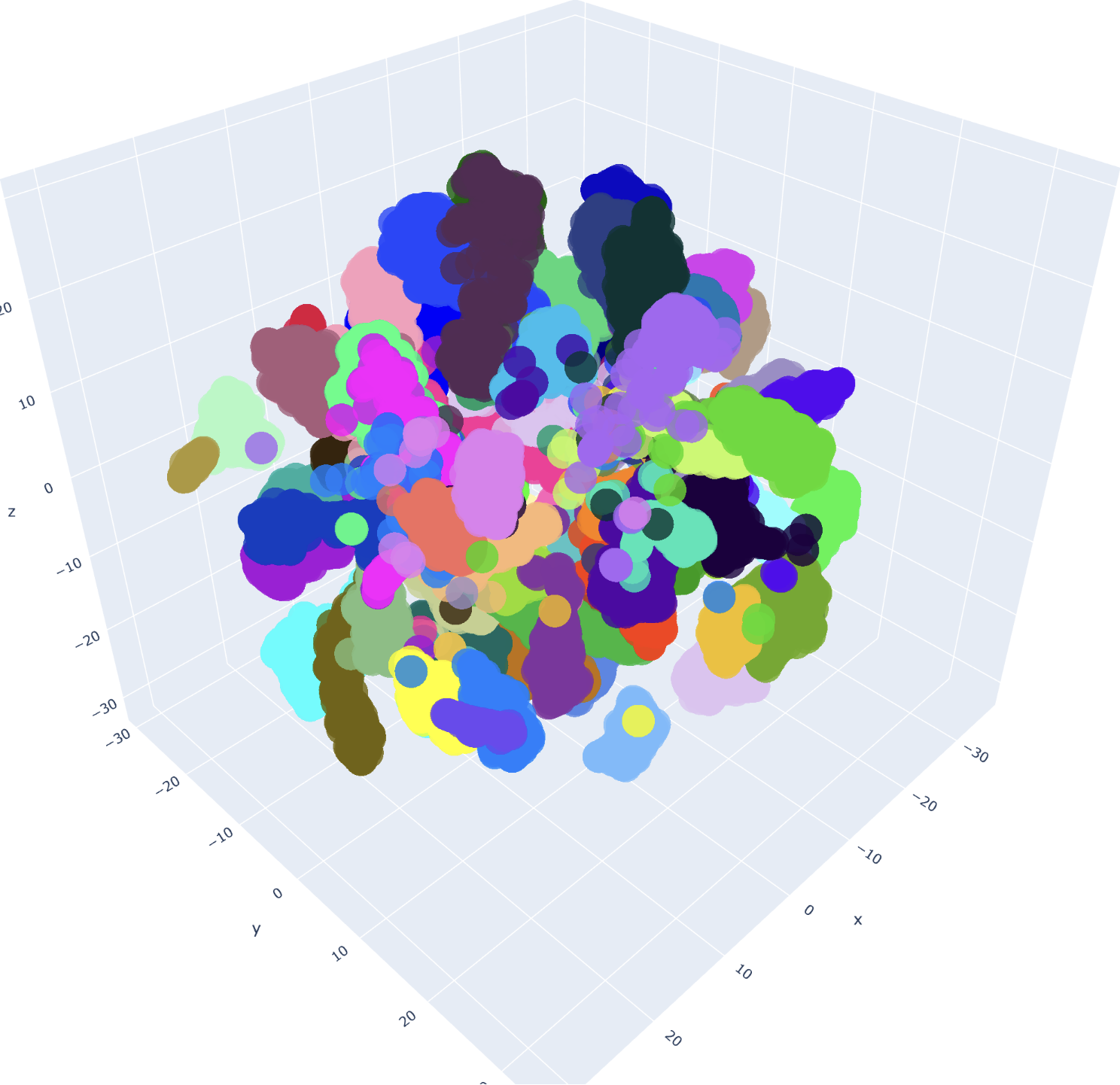}
        \subcaption{Fine logits $\boldsymbol{\gamma}_{\text{fine}}$}
    \end{subfigure}
    \caption{t-SNE of classifier outputs on InterHand2.6M test set.
    Coarse and fine logits exhibit different cluster granularities,
    consistent with the coarse-to-fine supervision design.}
    \label{fig:tsne_logits}
\end{figure}

We analyze the effect of gesture pretraining via t-SNE on the test set.
As shown in \cref{fig:tsne_pool}, baseline HRNet features exhibit poor
inter-class separation, while pretrained features form clearer clusters,
suggesting improved encoder discriminability.
\cref{fig:tsne_logits} shows that the coarse and fine classifier outputs
exhibit different cluster granularities, consistent with the intended
coarse-to-fine supervision hierarchy.

\subsection{Single-Hand 3D Pose Estimation}
\label{sec:singlehand_exp}

Results are shown in \cref{tab:single_hand_exp_results}.

\begin{table}[t]
\centering
\caption{Single-hand results on InterHand2.6M test set (mm).
PT denotes gesture-aware pretraining; full model additionally injects gesture guidance tokens in Stage~2.}
\label{tab:single_hand_exp_results}
\begin{tabular}{lcc}
\toprule
Method & MPJPE $\downarrow$ & MPVPE $\downarrow$ \\
\midrule
HaMeR~\cite{pavlakos2024hamer}              & 16.11 & 15.37 \\
EANet~\cite{park2023extract}                &  4.94 &  5.26 \\
EANet (w/ gesture pretraining)              &  4.86 &  5.17 \\
\midrule
Ours (w/o PT)                                   &  5.00 &  5.35 \\
Ours (full)                                     & \textbf{4.84} & \textbf{5.19} \\
\bottomrule
\end{tabular}
\end{table}

\paragraph{Effect on our method.}
Adding gesture pretraining and gesture guidance reduces both MPJPE and MPVPE,
showing that gesture priors improve both feature learning in Stage~1 and
token refinement in Stage~2.

\paragraph{Cross-architecture transferability.}
Replacing EANet's ImageNet-pretrained ResNet-50 with our gesture-pretrained
backbone---with no other changes to EANet---consistently reduces error.
This suggests that the benefit of gesture-aware pretraining is architecture-agnostic,
making it a general-purpose component for single-hand pose estimation.

\begin{figure}[t]
\centering
\begin{subfigure}{0.95\linewidth}
    \centering
    \subcaption*{Image \hspace{0.22\linewidth} EANet \hspace{0.22\linewidth} Ours}
    \includegraphics[width=\linewidth]{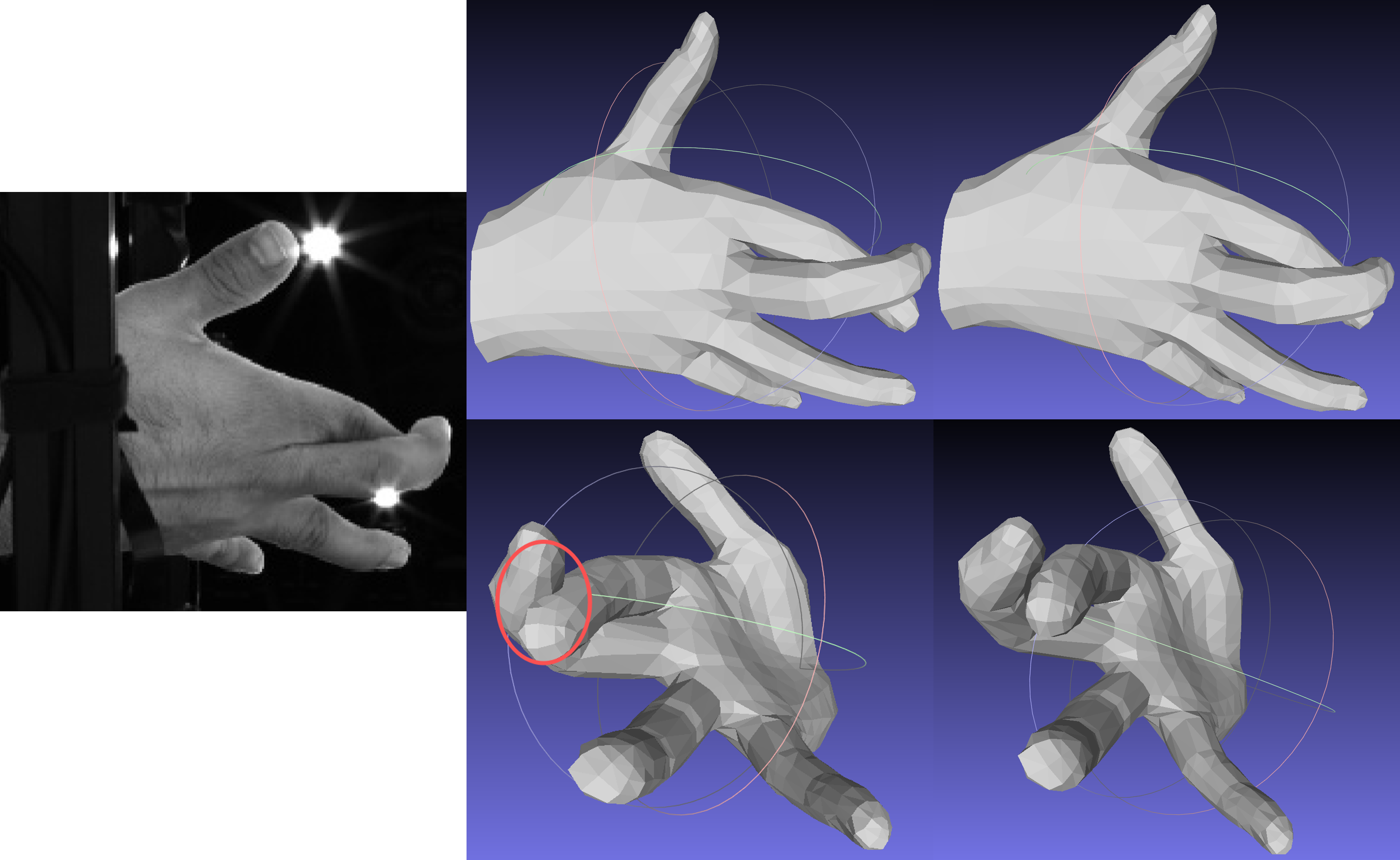}
    \caption{Hand-object occlusion.}
\end{subfigure}
\vspace{2mm}
\begin{subfigure}{0.95\linewidth}
    \centering
    \includegraphics[width=\linewidth]{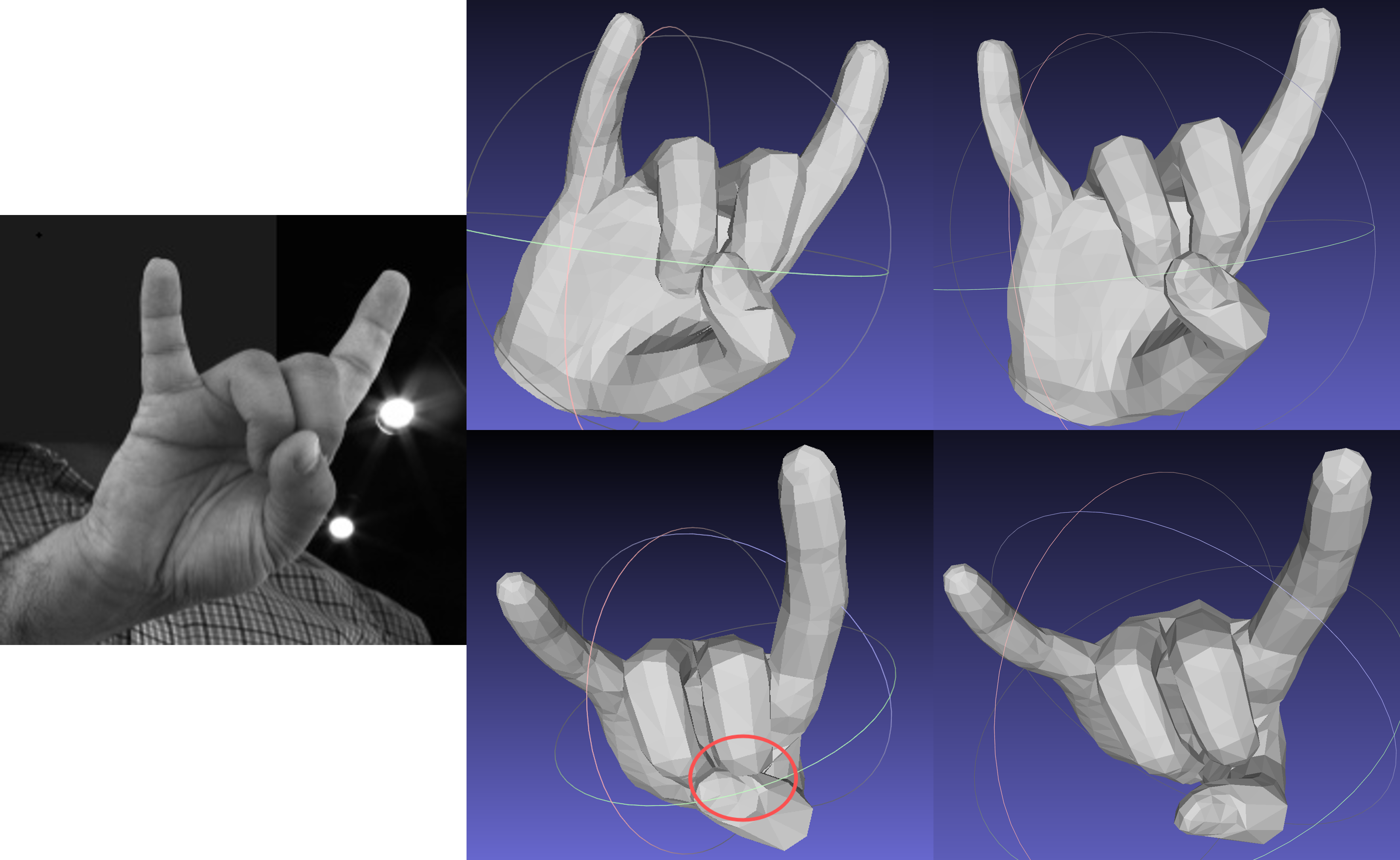}
    \caption{Finger self-occlusion.}
\end{subfigure}
\caption{Qualitative comparison under occlusion. Our method produces more
plausible hand meshes without interpenetration artifacts.}
\label{fig:mesh_comparison}
\end{figure}

\paragraph{Qualitative results.}
\cref{fig:mesh_comparison} compares EANet and our full method under occlusion.
In both examples, EANet produces mesh interpenetration artifacts (red ellipses),
while our method yields more anatomically plausible meshes, suggesting that
gesture-aware features help resolve ambiguous occluded poses.

\section{Conclusion}

We presented a gesture-aware pretraining strategy and a gesture-guided token fusion
framework for single-hand 3D pose estimation.
The key finding is that gesture semantics---learned from coarse-to-fine
hierarchical supervision on single-hand images---provide a useful and transferable
inductive bias: they improve accuracy in our own architecture and transfer
directly to EANet without any modification, suggesting their general utility.
Our method also produces more anatomically plausible meshes under occlusion,
attributed to the semantic constraints imposed by gesture pretraining.
Extending gesture-aware pretraining to two-hand and in-the-wild settings remains
a promising direction for future work.

{\small
\bibliographystyle{ieeenat_fullname}
\bibliography{egbib}
}

\end{document}